\documentclass{article}

\usepackage{PRIMEarxiv}

\usepackage[utf8]{inputenc} % allow utf-8 input
\usepackage[T1]{fontenc}    % use 8-bit T1 fonts
\usepackage{hyperref}       % hyperlinks
\usepackage{url}            % simple URL typesetting
\usepackage{booktabs}       % professional-quality tables
\usepackage{amsfonts}       % blackboard math symbols
\usepackage{nicefrac}       % compact symbols for 1/2, etc.
\usepackage{microtype}      % microtypography
\usepackage{fancyhdr}       % header
\usepackage{graphicx}       % graphics
\graphicspath{{media/}}     % organize your images and other figures under media/ folder
\usepackage{adjustbox}
\usepackage{numprint}
\usepackage{booktabs}
\usepackage{multirow}
\usepackage{tabu}
\usepackage[dvipsnames]{xcolor}
\usepackage{subcaption}
\usepackage{colortbl}

\title{Neural Architecture Transfer 2: \\A Paradigm for Improving Efficiency in Multi-Objective Neural Architecture Search
}

\author{
  Simone Sarti, Eugenio Lomurno, Matteo Matteucci \\
  Politecnico di Milano \\
  Department of Electronics, Information and Bioengineering\\
  Via Ponzio 34/5, 20133 Milan, Italy\\
  \texttt{\{simone.sarti, eugenio.lomurno, matteo.matteucci\}@polimi.it} \\
}

\begin{document}
\maketitle

\begin{abstract}
Deep learning is increasingly impacting various aspects of contemporary society. Artificial neural networks have emerged as the dominant models for solving an expanding range of tasks. The introduction of Neural Architecture Search (NAS) techniques, which enable the automatic design of task-optimal networks, has led to remarkable advances. However, the NAS process is typically associated with long execution times and significant computational resource requirements. Once-For-All (OFA) and its successor, Once-For-All-2 (OFAv2), have been developed to mitigate these challenges. While maintaining exceptional performance and eliminating the need for retraining, they aim to build a single super-network model capable of directly extracting sub-networks satisfying different constraints.
Neural Architecture Transfer (NAT) was developed to maximise the effectiveness of extracting sub-networks from a super-network.
In this paper, we present NATv2, an extension of NAT that improves multi-objective search algorithms applied to dynamic super-network architectures. NATv2 achieves qualitative improvements in the extractable sub-networks by exploiting the improved super-networks generated by OFAv2 and incorporating new policies for initialisation, pre-processing and updating its networks archive. In addition, a post-processing pipeline based on fine-tuning is introduced.
Experimental results show that NATv2 successfully improves NAT and is highly recommended for investigating high-performance architectures with a minimal number of parameters.
\end{abstract}

% keywords can be removed

\section{Introduction}\label{sec:introduction}

Deep learning has emerged as a significant revolution in recent years, significantly impacting various aspects of modern society. It has notably transformed numerous activities by leveraging artificial neural networks. These networks possess remarkable capabilities, outperforming conventional approaches in multiple tasks. One notable advantage is their ability to eliminate the requirement for manual feature engineering, as they autonomously discern meaningful patterns from the provided data. The effectiveness of deep learning networks stems from their meticulously designed layered architecture, enabling proficient feature extraction.
While human research endeavors have achieved notable advancements in performance, the resulting models have exhibited a trend towards increased size~\cite{liu2022convnet}. Consequently, their production necessitates not only specialized expertise but also hardware, energy, and production times that have become progressively unattainable~\cite{desislavov2021compute}.

Neural Architecture Search (NAS) emerged to address the need for innovative neural architectures that are universally applicable and don't require extensive expertise. Its primary goal is to automatically discover the optimal configuration for a given dataset and task~\cite{zoph2018learning}.
Over time, NAS has also incorporated considerations for computational and temporal constraints. These techniques aim to improve the performance of found models trading-off with the whole search process complexity. This includes minimizing time, energy consumption, and CO$_2$ emissions, as well as achieving a favorable trade-off between model performance and complexity in terms of parameters and operations.
Additionally, NAS must account for scenarios where devices running these models have limited memory or constraints. Therefore, it is crucial to broaden the scope of NAS beyond benchmark rankings and consider it as a means to address such limitations~\cite{liu2018darts}.

\begin{figure}[!t]
\centering
\includegraphics[width=0.6\textwidth]{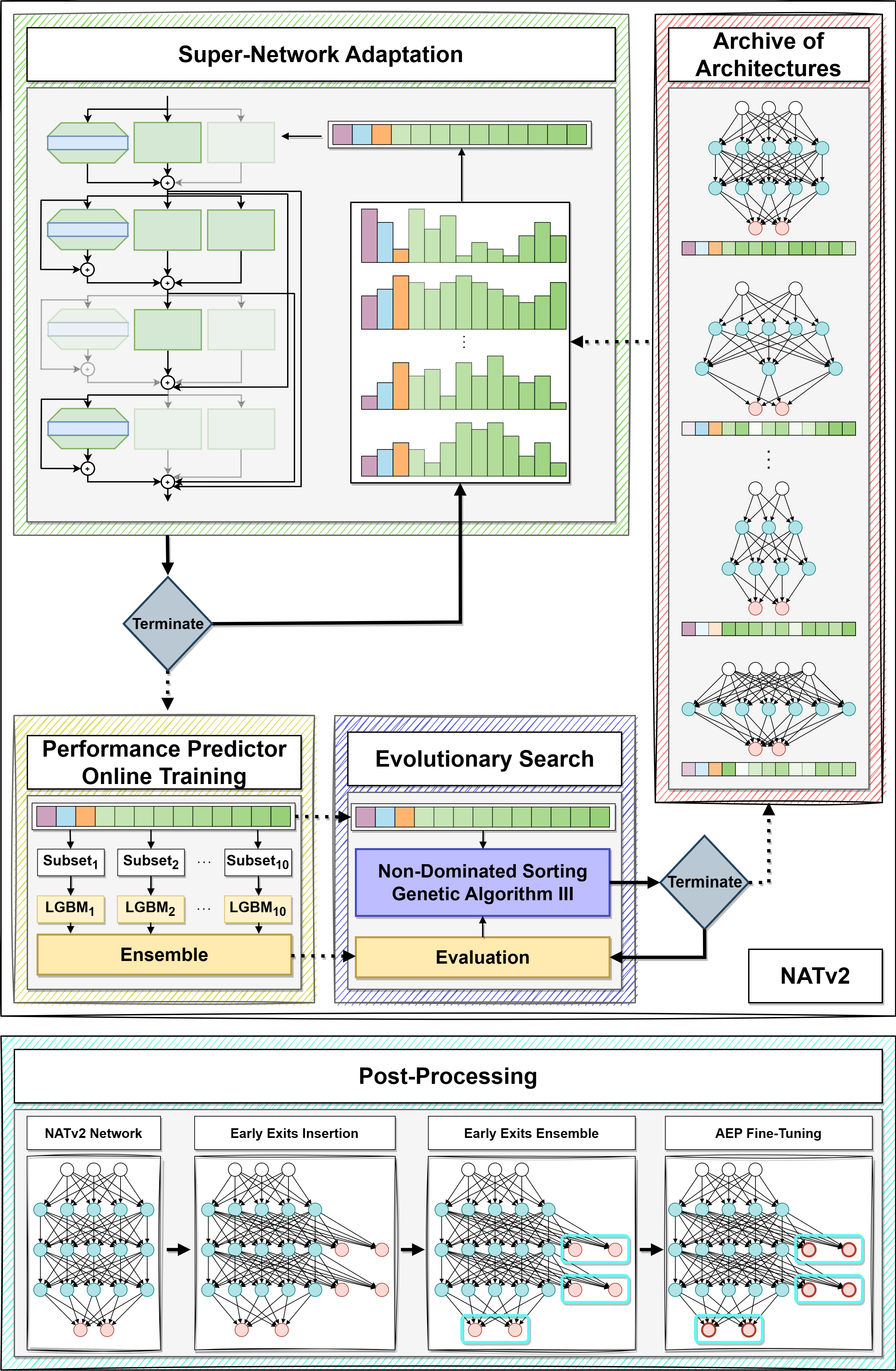}
\caption{The NATv2 summary diagram. The proposed algorithm designs customised architectures from a very large search space of possible state-of-the-art configurations. Multi-objective optimisation extends the work of NAT with new encoding and super-networks management techniques. New predictors provide accurate estimates for efficient evolutionary search. Once the optimal sub-network has been extracted, it is further refined by an additional post-processing step for fine-tuning.}
\label{fig:nat_process}
\end{figure}

The work known as Once-For-All (OFA) represents a significant milestone in this direction. As the name suggests, the objective of this approach is to perform massive computations in a single instance by constructing a super-network, from which sub-networks satisfying different constraints can be readily extracted, while maintaining excellent performance~\cite{cai2019once}.
This work has been successfully expanded through the technique known as Once-For-All-2 (OFAv2). The authors maintained the same underlying training principle as the original algorithm but adapted it to a network search space which was extended with proven techniques from the field of artificial neural network design, thereby elevating the obtained super-network to higher levels~\cite{sarti2023enhancing}.
The extraction of sub-networks represents the final and crucial step. While OFA already proposed a potential solution in this regard, the Neural Architecture Transfer (NAT) algorithm was specifically developed to maximize the effectiveness of this step.
NAT seeks to generate neural architectures that exhibit strong performance across diverse objectives by leveraging knowledge transfer and adaptation from pre-trained super-network models. It employs a combination of transfer learning and many-objective evolutionary search steps. Specifically, it adapts only the portions of the super-network corresponding to sub-networks discovered along the trade-off front by the search algorithm~\cite{lu2021neural}.

This paper introduces NATv2, an extension of NAT that enhances the capabilities of multi-objective search algorithms on dynamic super-network architectures. NATv2 replaces the original super-network, OFAMobileNetV3, used in NAT and pre-trained with OFA's Progressive Shrinking algorithm, with super-networks generated by OFAv2. Consequently, significant qualitative improvements are achieved in the extractable sub-networks' topology, allowing for the inclusion of parallel blocks, dense skip connections, and early exits.
To enhance the NATv2 archive, new policies are implemented for initialization, pre-processing, and updates. Moreover, a novel encoding type is proposed to accommodate these improvements, while the pipeline's predictors are upgraded to higher-performance techniques. Additionally, a post-processing pipeline based on fine-tuning is introduced, which further enhances model performance at a marginal increase in parameters and MACs.
By integrating all these advancements, NATv2 demonstrates the ability to generate image classification networks that surpass the accuracy achieved by NAT. Furthermore, NATv2 achieves this improvement with a reduced number of parameters and MACs. An overview of the proposed technique is depicted in Figure~\ref{fig:nat_process}.

The rest of the paper is divided into the following sections.
Section~\ref{sec:related} provides an introduction to NAS, the main works in the field of image classification, and the rationale behind their design choices.
Section~\ref{sec:method} describes the NATv2 method in detail, reporting on its workflow and paying special attention to the additions and improvements introduced by the new version.
The experiments performed, the configurations used and the qualitative and quantitative comparisons are described in Section~\ref{sec:results}.
Finally, Section~\ref{sec:conclusion} summarises the contributions of this work and concludes the manuscript.

\section{Related Works}\label{sec:related}

Neural Architecture Search (NAS) is an evolving research area within the deep learning community that combines various techniques from machine learning and optimization domains. The primary goal of NAS is to automatically design complex neural network architectures in an end-to-end process without human intervention. Despite its popularity in the AI community, NAS lacks standardised approaches due to the variety of techniques involved.
Howerver, Elsken \textit{et al.} proposed a widely accepted classification of NAS algorithms based on three key characteristics~\cite{elsken2019neural}:
\begin{itemize}
\item The \textit{search space}, referred to as the set of all possible architectures that can be found by the algorithm.
\item The \textit{search strategy}, which defines how the algorithm explores the search space to find optimal architectures for the given task.
\item The \textit{performance evaluation strategy}, which determines how to efficiently evaluate the quality of the architectures during the search process.
\end{itemize}

Early NAS research achieved remarkable model quality, but required significant computational resources. For example, NASNet, a pioneering work in the cell-based NAS approach, emerged as a competitive solution for image classification, rivalling state-of-the-art human-designed neural networks~\cite{zoph2018learning}.
Inspired by highly successful models such as ResNet~\cite{he2016deep} and InceptionNet~\cite{szegedy2015going}, which featured the sequential repetition of convolutional modules, NASNet aimed to identify the most effective set of layers and connections for the given task and encapsulate them in a computational macro unit called a cell. This cell could then be stacked multiple times according to the desired depth of the final network.
Unfortunately, this early work remained a challenge constrained within the confines of large computing centres, limiting its accessibility. However, a significant democratising advance came with the introduction of the PNAS algorithm.

PNAS introduced a sequential model-based optimisation technique into the NAS context to relax the computational and time constraints. The key idea is that the search process could start with thinner models and progressively add new parallel and sequential layers based on guidance from a surrogate model called predictor.
The predictor is a machine learning model that estimates the potential accuracy of each candidate architecture and dynamically adapts to the training results of previously sampled networks~\cite{liu2018progressive}.
In further advancements, the POPNAS series of algorithms were developed to improve the efficiency of the cell-based approach. These algorithms extended the use of predictors to estimate the training times of the architectures to be searched. This allowed a transition to multi-objective optimisation, which considered both minimisation of search times and quality of architectures by explicitly training networks on the Pareto front. This improvement significantly increased the search efficiency without compromising the quality of the best architectures obtained~\cite{lomurno2021pareto,falanti2022popnasv2,falanti2023popnasv3}.

An alternative approach is taken by works such as AmoebaNet~\cite{real2019regularized} and NSGANet~\cite{lu2019nsga}, which use evolutionary algorithms for architecture search and employ gradient descent techniques to optimise the weights of the discovered architectures.
In evolutionary algorithms, the search space represents the phenotype, while the architectures being searched are encoded as genotypes. At each iteration, a population of architectures is maintained, and their genotypes are modified by mutation and crossover operations to produce offspring.
Mutations in this context involve the random swapping of bits in the encoding, often resulting in the addition or removal of a layer, or the establishment or removal of a connection between two layers.
DARTS improves search efficiency by introducing a relaxation to the search space, making it continuous. This allows the entire search space to be represented as a single large network, known as a super-network, where each edge between layers is parameterised and part of the optimisation process.
This modification allows both model weights and structure to be optimised using gradient descent via two specialised training steps, one for the whole super-network and one for the sub-networks within it. In this way, the final architecture is nothing more than a sub-graph extracted from the super-network itself~\cite{liu2018darts}.
The efficiency of DARTS has contributed to the emergence of a new class of methods known as one-shot architecture searches. During the search process, each sampled architecture can be viewed as a composition of paths within the super-network. The weights of these paths are inherited from the super-network and fine-tuned to quickly assess the accuracy of the network~\cite{bender2018understanding,guo2020single}.
However, training super-networks is challenging as they are more sensitive to hyperparameter changes and weight co-adaptation. Specialised training techniques are required to avoid favouring only certain subsets of architectures in the final estimation phase.

\subsection*{Once-For-All}

Once-For-All (OFA) is a pivotal work in the realm of super-network-based NAS techniques, designed to maximize search efficiency. As a prominent component of hardware-aware NAS techniques, OFA has gained significant recognition and widespread adoption due to its Progressive Shrinking (PS) optimization strategy~\cite{cai2019once}.
This strategy not only enables the acquisition of excellent starting points for sub-network extraction but also concentrates the computational load into a single end-to-end training process. The PS algorithm is organized into four elastic steps, each comprising multiple phases. The first step, Elastic Resolution, involves randomly varying the size of input images. The second step, Elastic Kernel Size, gradually reduces the maximum kernel size for convolutional operators across the entire network. The third step, Elastic Depth, progressively decreases the minimum depth achievable for sub-networks. Finally, the fourth step, Elastic Width, aims to reduce the number of filters available for each convolutional layer.

The algorithm begins by defining the maximal network, which includes all PS parameters set to their maximum values. Subsequently, the PS training steps and phases are executed sequentially. The values unlocked by each phase for a specific elastic step remain available for selection in all subsequent training phases, enabling the addition of smaller networks to the search space. During each batch of images, a certain number of sub-networks are sampled from the current sample space and activated within the super-network. Their gradients are accumulated, and a single weight update step is performed.

Throughout the training process, the maximal sub-network serves as the teacher network for Knowledge Distillation~\cite{hinton2015distilling}. KD involves transferring knowledge from a large pre-trained network (referred to as the teacher network, in this case, the maximal sub-network) to a smaller network (referred to as the student network, in this case, an active sub-network). This is achieved by using the output of the teacher network as a soft target for the student network during training. By learning from the predictions of the teacher network, the student network can achieve similar performance to the larger teacher network while being smaller in size.
Once the PS algorithm is concluded, it is possible to sample sub-networks to extract their configuration via encoding, train surrogate models also known as performance predictors, and finally exploit them to identify the most suitable and performing sub-network according to different hardware constraints.

The results presented show how OFA effectively addresses the multi-model forgetting problem, which refers to the performance degradation caused by weight sharing when sequentially training multiple neural architectures within a super-network~\cite{benyahia2019overcoming}.
Among the notable macro-architectures introduced by the authors, the main one, called OFAMobileNetV3, is based on the Inverted Residual Bottleneck (IRB), i.e. an highly efficient type of block, based on depthwise separable convolutions, originally introduced in MobileNetV2~\cite{sandler2018mobilenetv2} and further refined in MobileNetV3~\cite{howard2019searching}.
By default, the network consists of five stages, each consisting of four blocks. In cases where the number or size of the feature maps from the input to the output of a block is changed, the residual connection cannot be used. Therefore, the IRB block is replaced by its sequential counterpart known as an Inverted Bottleneck (IB).

\subsection*{Neural Architecture Transfer}

Neural Architecture Transfer (NAT) is a recent advancement in NAS that builds on the OFA framework. It serves as an adaptive post-processing step that replaces the simple sub-networks extraction in the original OFA algorithm.
NAT aims to progressively transform a pre-trained generic super-network into a task-specific super-network. The goal is to directly search and extract sub-networks that achieve the best trade-off across a range of objectives from the task-specific super-network, without the need for re-training~\cite{lu2021neural}.

To optimise the efficiency of the super-network adaptation process, NAT selectively fine-tunes only those parts of the super-network that correspond to sub-networks whose structures can be directly sampled from the current trade-off front distribution. This approach saves computation by focusing on the parts that contribute to improvements in the current task.
The multi-objective evolutionary search in NAT is guided and accelerated by a performance prediction model that is updated online using only the best sub-networks configurations and their scores.  
This approach helps maintain a high-performance prediction model despite using a relatively small number of sub-networks as training samples.
By relying on the predictor, NAT can save time by avoiding evaluating each member of the intermediate populations generated during the evolutionary search.

Throughout the process, the best architectures found are gradually added to an archive. In the end, NAT returns three main outputs: the set of non-dominated sub-networks from the archive, which represent the best trade-offs across objectives; the set of high trade-off sub-networks, which are solutions where choosing any neighbouring solution would result in a significant loss in some objectives compared to a unitary gain in others; and the resulting task-specific super-network, which can be reused as a starting point for a new NAT process in different deployment scenarios.

\subsection*{Once-For-All-2}

Once-For-All-2 (OFAv2) represents a significant evolution from its original version, aiming at the same goal of a one-shot NAS algorithm to construct a super-network from which models suitable for different devices can be extracted. This new version introduces significant improvements, particularly in the search space~\cite{sarti2023enhancing}. In particular, the original OFAMobileNetV3 macro-architecture has been enhanced by the authors through the incorporation of parallel blocks, dense skip connections, and early exits, the latter taking advantage of the Anticipate Ensemble and Prune (AEP) technique ~\cite{sarti2023anticipate}. These additions increase the flexibility and performance of the super-network.

To accommodate the aforementioned architectural changes, the PS training algorithm has been extended to the Extended Progressive Shrinking (EPS) algorithm, incorporating two new elastic steps: Elastic Level and Elastic Exit. Elastic Level supports parallel networks, while Elastic Exit is applied in the presence of early exits.
In addition, OFAv2 introduces a novel teacher network extraction strategy. This strategy dynamically updates the teacher network at the end of each EPS step, ensuring the transfer of relevant knowledge for subsequent training steps.

\section{Method}\label{sec:method}

\begin{figure*}[t]
\centering
\includegraphics[width=0.95\textwidth]{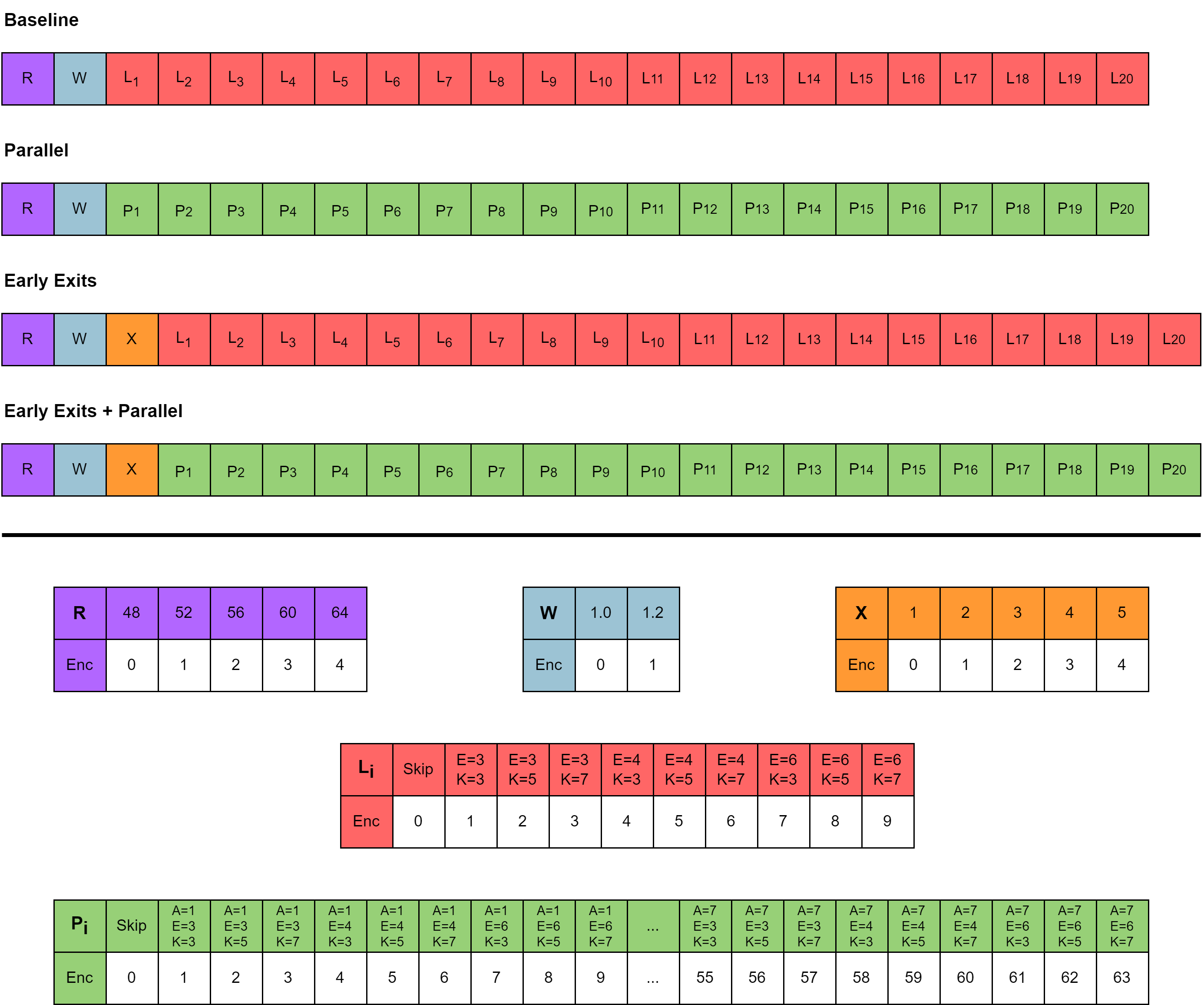}

\caption{
The encodings representing the possible sub-networks within different types of super-networks have the following structure. $R$ encodes the value corresponding to the size of the input images. $W$ encodes the value of the width multiplier, which determines the width of the network architecture. $X$ encodes information about the selected exit, specifically for super-networks that support early exits. $L_i$ encodes the configuration of the $i$\textsuperscript{th} IRB/IB block for non-parallel networks. $P_i$ encodes the configuration of the $i$\textsuperscript{th} level, i.e. set of blocks in parallel, for parallel networks. The ``Baseline" configuration represents the encoding used in NAT, while the other configurations represent the encodings proposed and used in NATv2.}
\label{fig:tinynat_search_spaces}
\end{figure*}

This section presents the second version of the Neural Architecture Transfer (NATv2) algorithm, which builds on the Once-For-All-2 (OFAv2) technique used to generate the the super-networks used as staring points. The focus is on the modifications made to the original algorithm to accommodate any architecture generated by OFAv2. In addition, changes to the sub-network sampling method and performance predictor are presented.
NATv2 introduces two new steps in its pipeline: a pre-processing step, which incorporates the new archive initialisation method, and a two-stage post-processing method.
The pre-processing step is designed to ensure that the archive is appropriately set up to start with a set of performing architectures from the beginning.
The post-processing step is applied after the conclusion of the algorithm to fine-tune and refine the selected architectures, ultimately improving their performance. 
Being an upgraded version, it is assumed that all steps and algorithms not explicitly mentioned have remained unchanged from the original approach.

\subsection*{Expanded Search Space}

NATv2 requires a new encoding paradigm to enable evolutionary search on OFAv2 super-networks. The existing encoding method used by NAT to represent sub-network structures in OFAMobileNetV3 is insufficient to capture the full range of architectural permutations. However, it serves as a starting point and undergoes precise modifications to meet the new set of constraints.
For the OFAMobileNetV3 architecture, the baseline NAT representation utilizes integer-encoded strings consisting of 22 values, as depicted in Figure~\ref{fig:tinynat_search_spaces} under the label ``Baseline". Each compressed representation contains specific information. The first value encodes the resolution $R$ of the input image for the network. The second value represents the active width multiplier $W$.
Width multipliers serve as scaling factors for the number of filters during the execution of the OFA algorithm. In the vanilla configuration, as in NATv2, two distinct super-networks are constructed, with $W$ values of 1.0 and 1.2, respectively. These two initial encoding rules are retained in the NATv2 encoding scheme.

\begin{figure}[!t]
\centering
\includegraphics[width=0.4\textwidth]{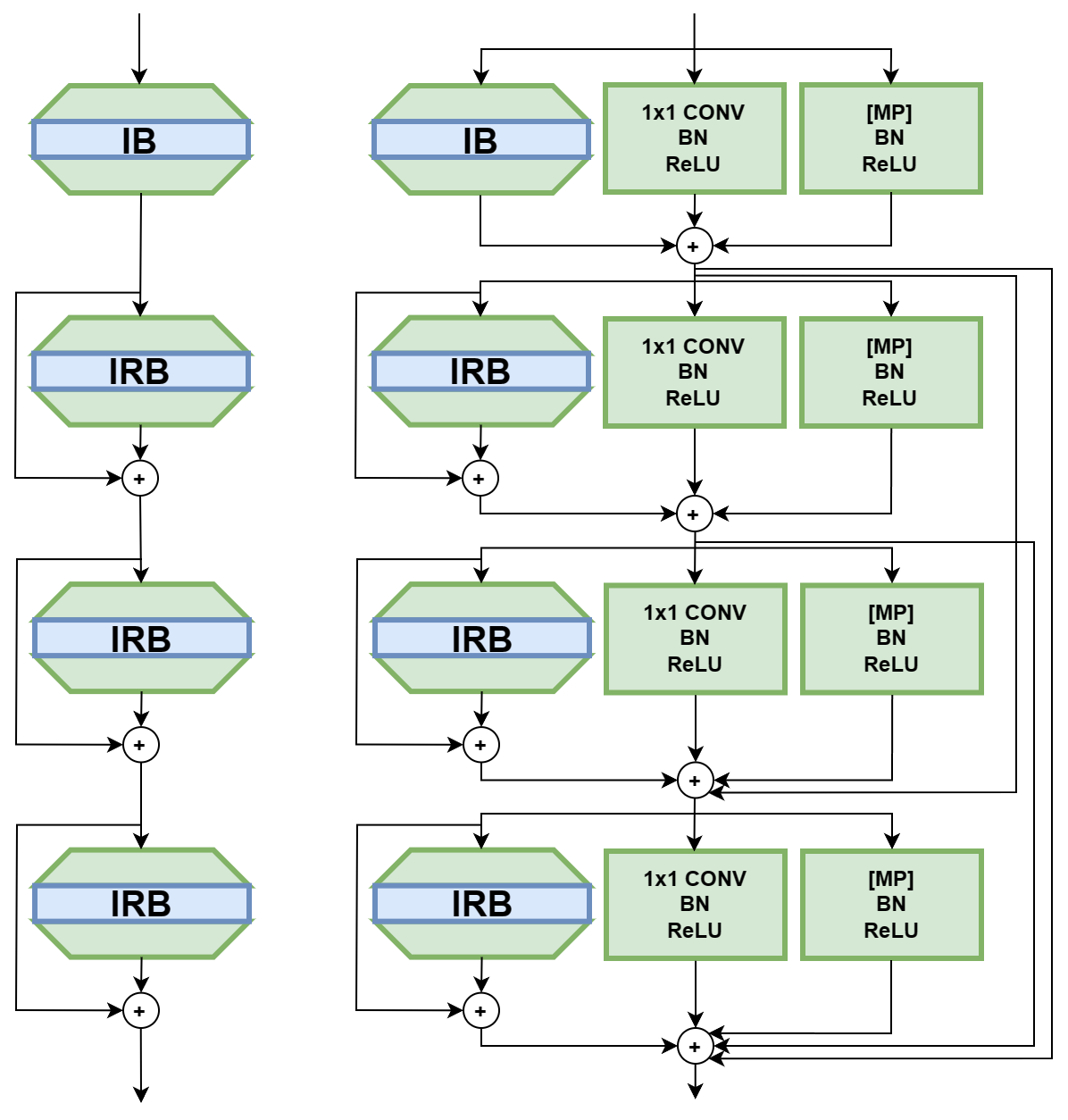}
\caption{In NATv2, the super-networks can be enhanced by the introduction of two new blocks running in parallel to the existing IRBs/IBs and dense skip connections within each stage. 
The first parallel block consists of a pointwise convolutional layer, a batch normalisation layer, and a non-linear activation function. The second parallel block consists of a batch normalisation layer followed by a non-linear activation function and eventually preceded by a max pooling operator. These blocks represent a lighter alternative to IRBs/IBs which contributes to the diversification of possible sub-network topologies and has the potential to improve computational efficiency.}
\label{fig:OFAv2pb}
\end{figure}

\begin{figure}[!t]
\centering
\includegraphics[width=0.3\textwidth]{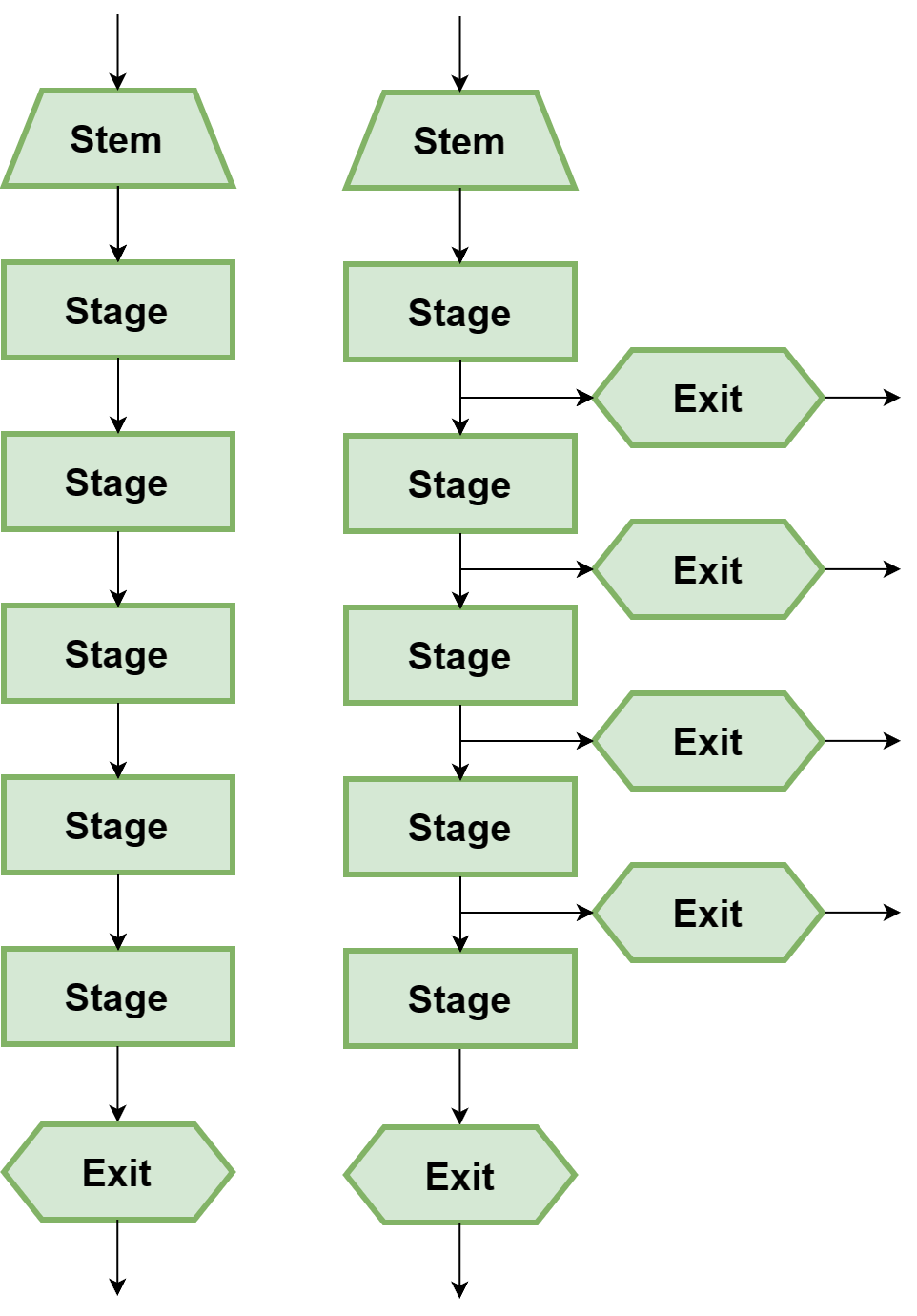}
\caption{The NATv2 super-networks can contain several intermediate outputs, called early exits, which are strategically placed after each network stage. The idea behind early exits is that intermediate outputs may have comparable or better performance than the final one.
Each early exit serves as an individual output that can be used independently for prediction. Alternatively, the outputs of multiple early exits can be aggregated or combined using an ensemble technique.}
\label{fig:OFAv2ee}
\end{figure}

In NAT, the set of 20 encoded values represents the combinations of kernel size $K$ and expansion ratio $E$ for each of the 20 internal IRB or IB blocks. To allow the inclusion of parallel blocks in the super-networks, as illustrated in Figure~\ref{fig:OFAv2pb}, these pairs are expanded into triplets by introducing the additional term $A$. The value of $A$ ranges from 1 to 7, covering all possible permutations of parallel blocks activation states. Including the special value 0, which represents the i\textsuperscript{th} block or level being excluded from the network, thus reducing the stage depth, each of the 20 P$_i$ values can take up to 64 different values. This expansion of the search space allows greater exploration without changing the encoding size, as shown in the second row of Figure~\ref{fig:tinynat_search_spaces} labelled ``Parallel".

In order to incorporate early exits into super-networks, and to capture sub-networks capable of making inferences at intermediate stages, it is necessary to encode this information appropriately. This ensures that the extracted sub-networks have the potential to make inferences at the end of each stage of the super-network, as illustrated in Figure~\ref{fig:OFAv2ee}.
To allow for this, a new variable called $X$ is introduced in the third position of the encoded strings, as shown in the third row of the Figure~\ref{fig:tinynat_search_spaces} labelled ``Early Exits". The value of $X$ corresponds directly to the index of the selected exit, providing information about the selected exit if early exits are present in the super-network architecture. The final encoding used in NATv2 is shown in the fourth line of Figure~\ref{fig:tinynat_search_spaces} and is named ``Early Exits + Parallel". It encompasses the combined modifications necessary to incorporate parallel blocks and early exits. Compared to the version employed in NAT, the search space has been significantly increased at the minimal cost of one additional character in the encoding string.

\subsection*{Archive Initialization and Update}

NATv2 takes a different approach to managing the archive of optimal sub-networks. The changes primarily affect two key stages of the NAT algorithm: the archive initialisation and archive growth steps.
With respect to the archive initialisation step, NAT directly samples architectures from the search space, thus evenly distributing possible values within the encoding. As a result, the initial archive has a strong bias towards networks with a maximum stage depth of 4. This bias arises from the fact that skippable IRB blocks (specifically, the 3\textsuperscript{rd} or 4\textsuperscript{th} block within a stage) can be encoded with values ranging from 0 to 9, but only assigning a value of 0 will cause these blocks to be skipped. Essentially, due to uniform sampling, all values have an equal chance of being selected, making stages with a depth of 3 uncommon and stages with a depth of 2 quite rare. This problem becomes more pronounced when parallel blocks are supported, as the number of possible encodings for each network level is expanded up to 64 values.
To address this issue, NATv2 approaches the archive initialisation phase by sampling subnets in a way that ensures uniformity not within the search space domain, but rather within the depth (for both stages and networks) and in the block configurations (for both parallel and non-parallel levels). This adaptation encourages greater heterogeneity in the NATv2 initialisation process, while improving the generalisation of the predictor through a more diverse training set.

In contrast to the archive growth step in NAT, NATv2 introduces a sub-network replacement step. Instead of starting with a limited set of architectures and allowing the archive to grow iteratively by adding newly discovered sub-networks, NATv2 directly populates the archive with the maximum number of architectures right from the start.
During each iteration, the weakest architectures within the archive are replaced with those obtained through the evolutionary search. Consequently, the size of the archive remains constant throughout the process. This approach aims to enhance the average quality of the architectures within the archive, leading to improved performance of the predictor model. Notably, this improvement is particularly evident in the early iterations due to the larger pool of input data available for the predictor to learn from. The overall effect is an increased capability of the performance predictor in gauging the quality of sub-networks in NATv2.

In addition, NATv2 includes a pre-processing step within the archive initialization phase. This pre-processing entails sampling a significantly larger number of architectures, specifically ten times larger than the desired archive size, denoted as $A_s$. The sampled sub-networks are evaluated and compared, and only the top $A_s-2$ networks, along with the maximal and minimal networks, are selected to form the initial archive.
Introducing a set of high-quality architectures at the beginning of the algorithm can contribute to obtain better performing sub-networks throughout the process. It is worth noting that the inclusion of the pre-processing step comes at the cost of increased execution time, but this is only a one-time cost per execution. The extent of this impact depends on the size of the initially sampled architecture set. In NATv2, the archive size $A_s$ has been defined as 300.

\subsection*{Performance Predictor}

The execution of the evolutionary search process in NATv2 generates a significant number of sub-networks. However, evaluating the performance of each of these sub-networks individually would render the algorithm computationally impractical, even with the use of weight sharing.
To overcome this challenge, NATv2, like its predecessor, relies on a performance predictor model. This predictor performs a regression task and is trained online. At the start of each NAT iteration, the predictor model is fitted by taking as input the set of encodings corresponding to the sub-networks currently present in the archive, with their corresponding top-1 accuracy serving as the target.
By leveraging this dataset, the predictor model is trained to predict the performance of architectural encodings that it has not encountered before. This approach allows NATv2 to efficiently estimate the performance of numerous sub-networks without having to evaluate each one individually, making the algorithm computationally tractable.

In NATv2, a significant expansion of candidate predictor models has been undertaken. Initially, the following machine learning models were considered:

\begin{itemize}
\item Gaussian Process (GP)~\cite{williams1995gaussian},
\item Radial Basis Function (RBF)~\cite{bors2001introduction},
\item Multilayer Perceptron (MLP)~\cite{rumelhart1985learning},
\item Classification and Regression Tree (CART)~\cite{loh2011classification},
\item Radial Basis Function Ensemble (RBFE)~\cite{bors2001introduction}.
\end{itemize}
In order to thoroughly investigate the effectiveness of the predictor, various regression mechanisms have been explored, leading to the inclusion of the following models

\begin{itemize}
\item Support Vector Regressor (SVR)~\cite{drucker1996support},
\item Ridge Regressor~\cite{hoerl1970ridge},
\item K-Nearest Neighbours Regressor (KNN)~\cite{fix1989discriminatory},
\item Bayesian Ridge Regressor~\cite{tipping2001sparse}.
\end{itemize}
In addition, given their claimed success in the literature, the following models have been added to the candidate predictor list:

\begin{itemize}
\item End-to-End Random Forest-based Performance Predictor (E2EPP)~\cite{sun2019surrogate},
\item Light Gradient Boosting Machine (LGBM)~\cite{ke2017lightgbm},
\item Catboost~\cite{prokhorenkova2018catboost}.
\end{itemize}
This extensive selection of machine learning models allows a comprehensive exploration of potential predictors in NATv2, highlighting the importance of the role of the predictor in the algorithm.

\begin{figure}[t]
\centering
\includegraphics[width=0.45\textwidth]{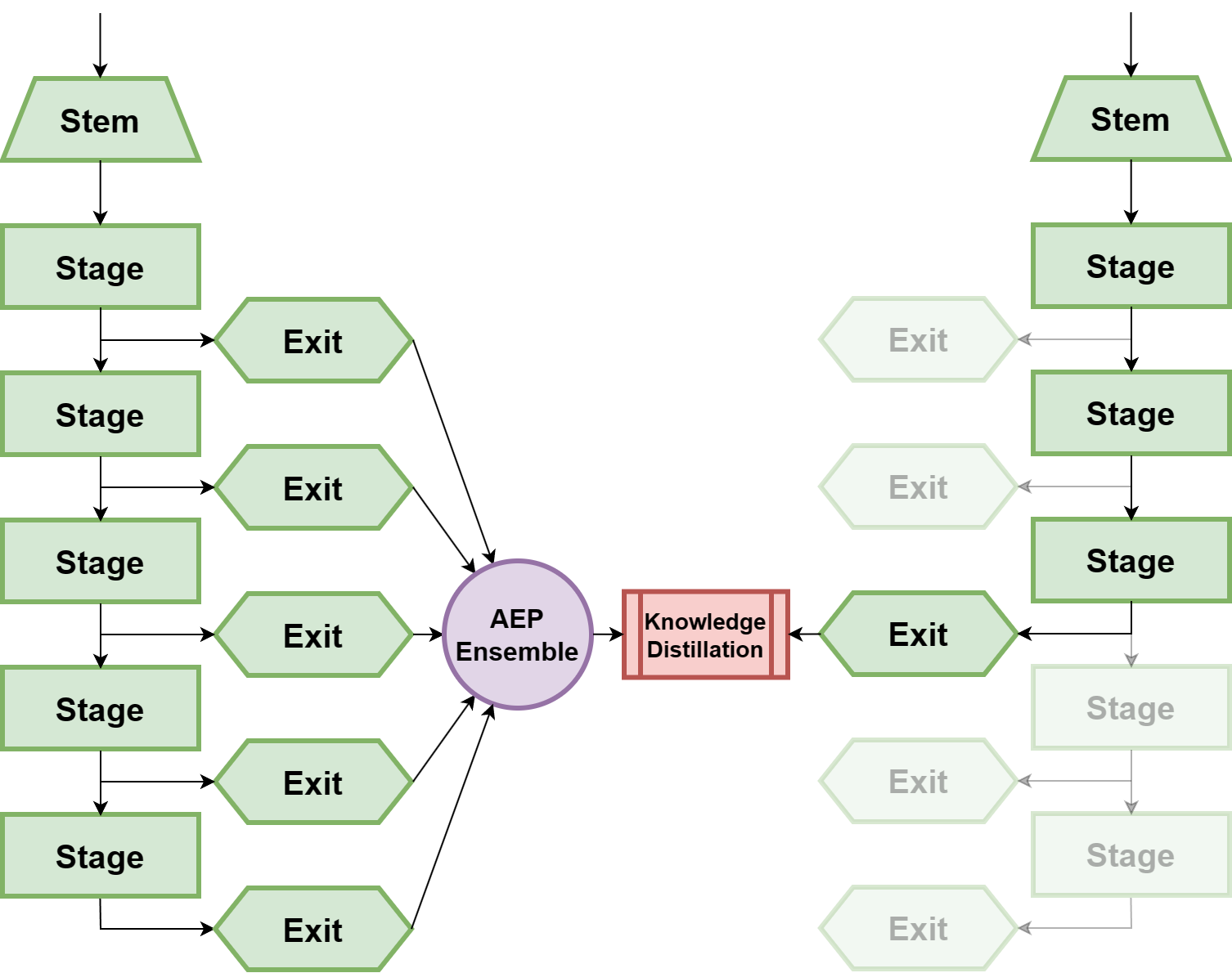}
\caption{The scheme of the Knowledge Distillation technique for networks with early exits called ENS-KD, presented in OFAv2 and used in NATv2.}
\label{fig:OFAv2kd}
\end{figure}

\subsection*{Training Networks with Early Exits }

Special attention was paid to the training algorithm applied to super-networks with early exits, enriching it with new techniques introduced and proven effective in both the AEP and OFAv2 works. 
In particular, the AEP training method is relied upon; given a network with multiple exits, the network undergoes a form of joint training by creating a weighted ensemble of its exits. Different weighting strategies can be used to adjust the contribution of each exit, but in general the benefits of this training method were clearly demonstrated in the original study.
Another important technique is ENS-KD, a new knowledge distillation technique introduced in OFAv2, which is also based on the AEP approach. 
As shown in Figure~\ref{fig:OFAv2kd}, given a teacher network with multiple exits, the knowledge transferred to the student isn't limited to that available at the last layer of the teacher network, but rather information from all its exits is weighted and combined according to the AEP method to distill more significant information, resulting in improved performance of the student networks.

Returning to the training of the NATv2 super-networks with early exits, during the warm-up phases, the maximal networks extracted from the OFAv2 super-network under experimentation, i.e. those corresponding to width multipliers 1.0 and 1.2, are trained using the AEP training method, following the DESC weighting strategy. 
The use of the AEP joint training method is possible because the maximal network within an OFAv2 super-network with early exits retains all the exits present in the super-network.
For the NATv2 adaptation phase, in which the super-network is fine-tuned by sequentially activating sub-networks within it, a training algorithm corresponding to the last phase of the EPS algorithm is used. 
This is because NAT immediately makes all the possible values of each elastic parameter available for sampling. On the contrary, the steps and phases of EPS progressively reveal new elastic parameters and their values. It is only in the last phase of EPS that all elastic parameter values are available for sampling.

NAT uses Knowledge Distillation to improve sub-networks during the super-network adaptation phase.
In NATv2, the standard Knowledge Distillation technique is replaced by the ENS-KD technique when performing the adaptation step on early exit super-networks. 
To be consistent with the EPS training that the super-networks underwent in OFAv2, the sub-networks activated during the super-network adaptation step in NATv2 are single exit networks.

\subsection*{Post-Processing}

The multi-objective optimization nature of the search process in NATv2 may result in the best sub-networks achieving high performance across multiple objectives but still not reaching their maximum classification potential.
To address this, two distinct fine-tuning post-processing methods are introduced to maximize the accuracy of these networks. The first method is applicable to sub-networks derived from any super-network, while the second method is specifically designed for sub-networks extracted from super-networks with early exits. Both methods consist of two sequential phases.

In the first phase, the optimal number of training epochs, denoted as $e$, is determined for the given sub-network through fine-tuning on the target dataset. During this phase, the performance of the sub-network is continuously evaluated on the corresponding validation set.
In the second phase, the sub-network is fine-tuned for $e$ epochs using the combination of the training and validation sets. Finally, the test classification performance is computed and returned.

The difference between the two post-processing methods lies in the fine-tuning algorithm used. The first method directly fine-tunes the networks returned by NATv2 as they are, i.e. single exit networks. On the other hand, the second post-processing method utilizes the Anticipate, Ensemble and Prune (AEP) technique~\cite{sarti2023anticipate}. In this method, all exits above the one selected by NATv2 for the sub-network are extracted from the super-network and reattached to the sub-network. A joint fine-tuning of the exits is then performed.
The second post-processing method allows for greater gains in accuracy compared to the traditional fine-tuning method. However, it comes at the cost of a slightly increased number of parameters and MACs.

\section{Results and Discussion}\label{sec:results}

This section outlines the experiments conducted to evaluate the performance of NATv2 and compare it to NAT, the method on which our approach is based. First, the experiments are described in detail, including the experimental setup, implementation specifics, and hardware used to ensure reproducibility. Then, three ablation studies are presented to provide insight into the intermediate steps and to evaluate the decision-making process that led to the final results. The first study focuses on identifying the optimal performance predictor to use during the search. The second examines the effectiveness of the new super-networks and their encodings compared to those used in NAT. The last explores different post-processing optimisation strategies to determine the most effective approach.
Finally, we present the results of our study by comparing the performance of NAT and NATv2, with and without post-processing, on the proposed datasets and configurations. The aim is to provide compelling evidence to support our claims and convince the reader of the superiority of NATv2.

\subsection*{Experiments Configuration}

The experiments conducted for NATv2 used the super-network obtained with OFAv2 using the Extended Progressive Shrinking (EPS) technique. In contrast, the NAT experiments used the baseline OFAMobileNetV3 super-network obtained by OFA. Two versions of the same pre-trained super-networks were required to allow for the warm-up steps in both algorithms. The width multiplier configurations used were $W=1.0$ and $W=1.2$, maintaining consistency with those used in the NAT paper.

While the OFA networks were trained on ImageNet, the OFAv2 networks were trained on the Tiny ImageNet dataset instead. To maintain consistency, and due to resource constraints, the Tiny ImageNet dataset was also used instead of ImageNet in both the NAT and NATv2 configurations. The experiments were performed on three widely used image classification datasets: CIFAR-10, CIFAR-100 and Tiny ImageNet. Further details on these datasets are given in Table~\ref{table:natv2_datasets}.

To ensure comparable results, the same set of hyperparameters was used for all experiments. The NATv2 models were trained using the SGD optimiser with a momentum of 0.9 and a weight decay parameter set to $3\cdot10^{-4}$. The learning rate was initially set to $2.5\cdot10^{-3}$ and adjusted using a cosine annealing scheduler. A batch size of 256 was used, and during each super-network adaptation epoch, four sub-networks per batch were sampled. The same hyperparameters were used for the warm-up phases, with the exception of the initial learning rate, which was set to $7.5\cdot10^{-3}$.

\begin{table}[t]
\centering
\caption{The details of the datasets used in this work in terms of number of classes and splits.}
\resizebox{0.675\textwidth}{!}{%
\begin{tabular}{|lcccc|}

\hline
\textbf{Dataset}                                    & \textbf{Classes}   & \textbf{Train size} & \textbf{Validation size}    & \textbf{Test size} \\ 
\hline 
Tiny ImageNet~\cite{le2015tiny}                     & \numprint{200}        & \numprint{85000}    & \numprint{15000}            & \numprint{10000}    \\ 
CIFAR10~\cite{krizhevsky2009learning}              & \numprint{10}         & \numprint{45000}    & \numprint{5000}             & \numprint{10000}    \\ 
CIFAR100~\cite{krizhevsky2009learning}             & \numprint{100}        & \numprint{45000}    & \numprint{5000}             & \numprint{10000}    \\ \hline

\end{tabular}}%
\label{table:natv2_datasets}
\end{table}

In order to maximise the effectiveness of the post-processing step, numerous combinations of optimisers and learning rates were tested. These experiments aimed to determine the optimal configurations and were conducted on sub-networks derived from the initial NATv2 experiments using the CIFAR-100 dataset, with NATv2 run with the objectives ``Accuracy \& Params" and ``Accuracy \& MACs".
To explore different post-processing combinations for single exit networks and early exit networks, the following optimisers were tested: SGD, AdamW~\cite{loshchilov2017decoupled}, and Ranger (the combination of the LookAhead~\cite{zhang2019lookahead} and RAdam~\cite{liu2019variance} optimisers). For each optimiser, two initial learning rates were tuned, i.e. $10^{-4}$ and $10^{-5}$.

In the case of AEP-based post-processing, which is applicable to sub-networks derived from early exit super-networks, experiments were conducted for all four AEP exit weighting strategies. When the SGD optimiser was used for post-processing, the previously reported values for momentum and weight decay were used, while no additional hyperparameters were specified for the other two optimisers.
The batch size for these experiments was set to 64, and the networks were trained for a maximum of 150 epochs using a cosine annealing learning rate scheduler. Early termination was used, with a patience value of 30 epochs based on validation loss.
All models were implemented using PyTorch 1.12.1 and experiments were run on an NVIDIA Quadro RTX 6000 GPU. The evolutionary algorithms used for the NAT and NATv2 search steps, specifically NSGA3~\cite{blank2019investigating}, were obtained from the pymoo library~\cite{blank2020pymoo}.

\begin{figure*}[!t]
\centering
\includegraphics[width=\textwidth]{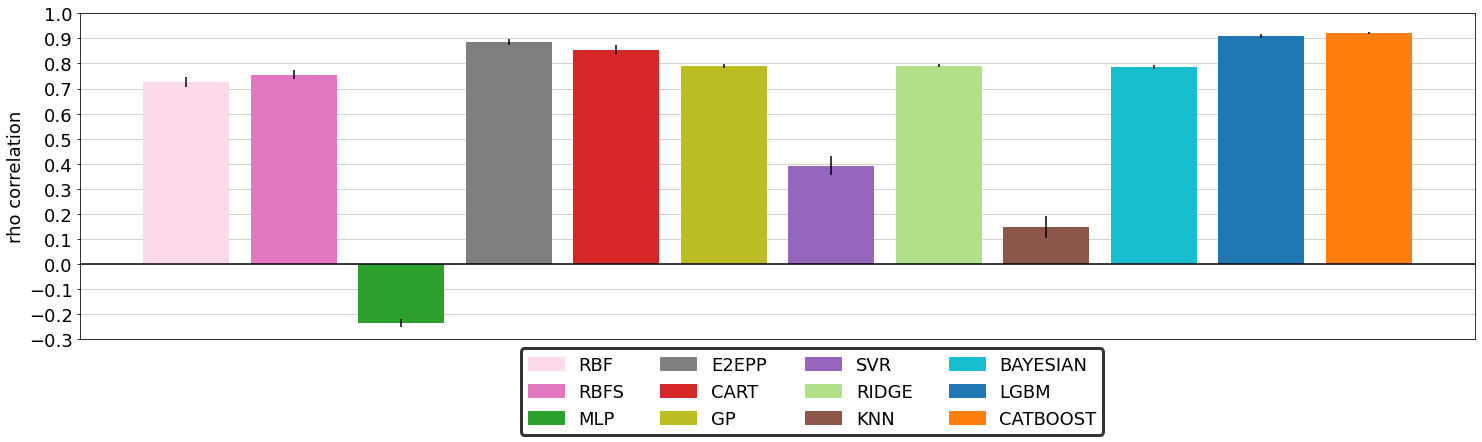}
\caption{The rho correlation values for different models of performance predictors considered in this paper. The predictors were trained on 300 samples using the proposed integer encoding. For a list of predictor names and their acronyms, see Section~\ref{sec:method}.}
\label{fig:predictor_by_network}
\end{figure*}

\begin{figure*}[!t]
\centering
\includegraphics[width=\textwidth]{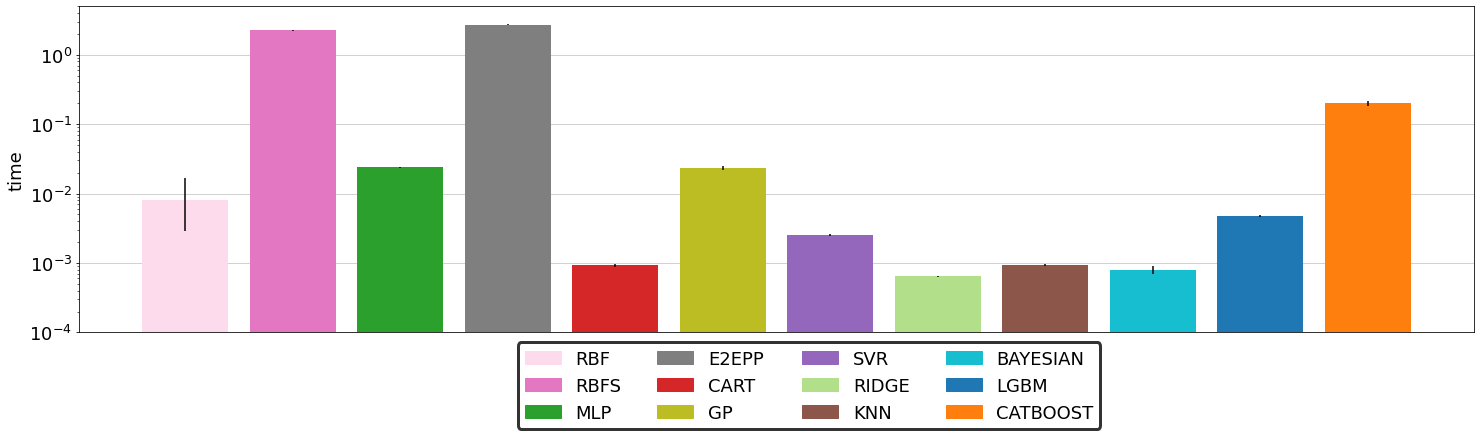}
\caption{The time values for different models of performance predictors considered in this paper. The predictors were trained on 300 samples using the proposed integer encoding. The y-axis is represented in logarithmic form. For a list of predictor names and their acronyms, see Section~\ref{sec:method}.}
\label{fig:time_predictor_by_network}
\end{figure*}

\begin{figure*}[!t]
\centering
\includegraphics[width=\textwidth]{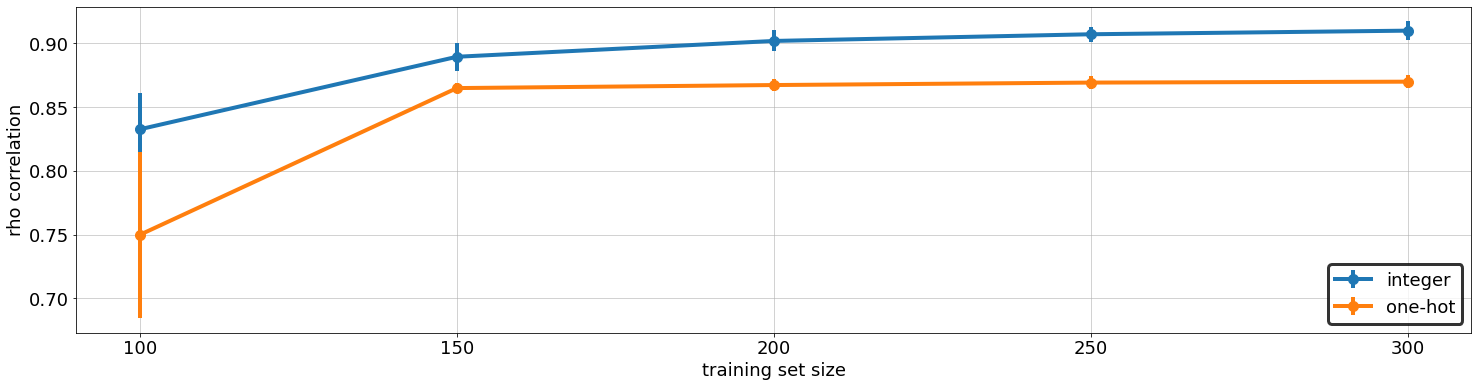}
\caption{The rho correlation values achieved by the LGBM predictor for different training set sizes and encodings.}
\label{fig:predictor_by_encoding}
\end{figure*}

\begin{figure*}[t]
\centering
\includegraphics[scale=0.65]{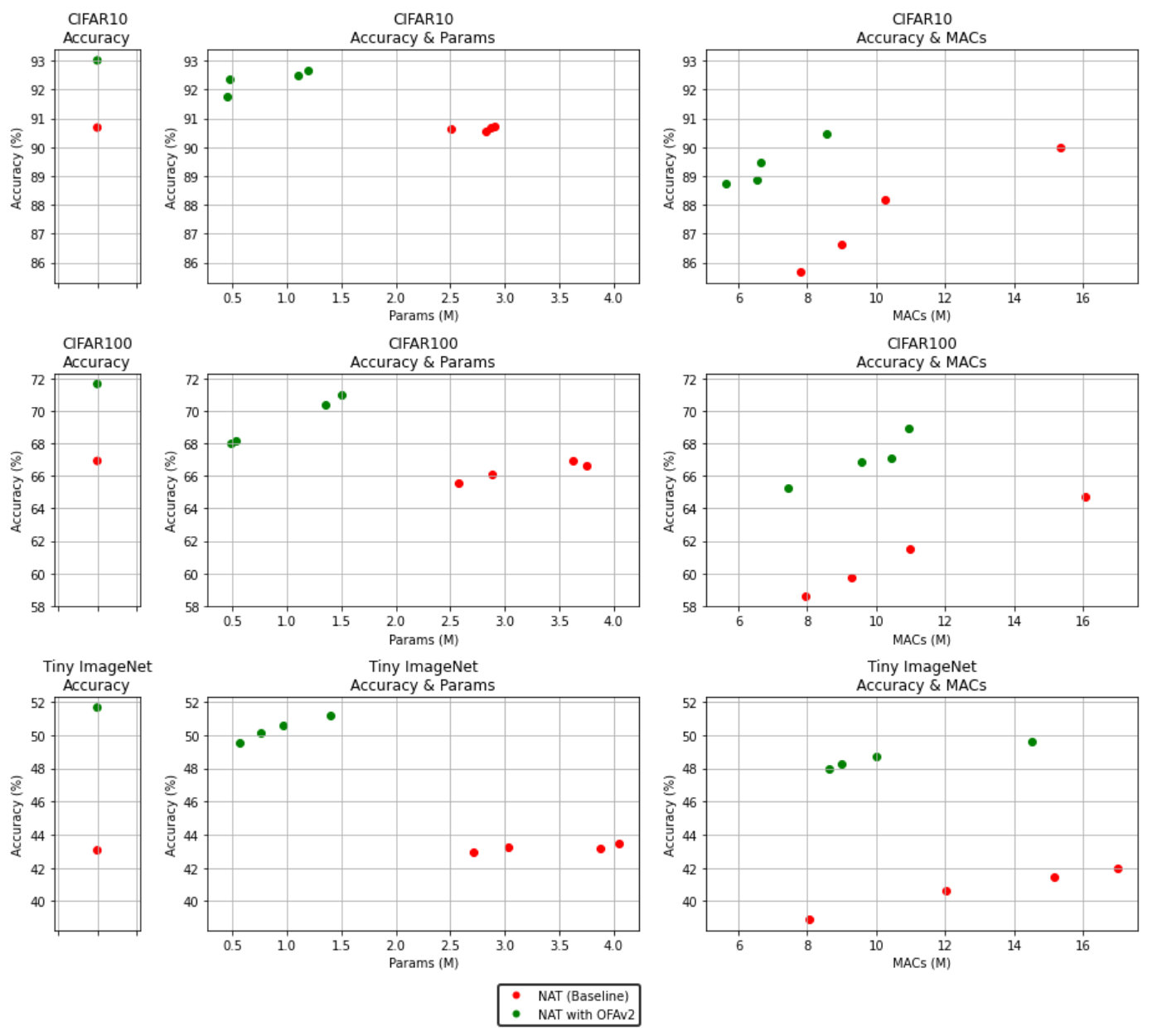}
\caption{The results of the study of the effectiveness of the OFAv2 super-networks compared to the OFA super-network within the NAT algorithm, based on the proposed datasets and optimisation strategies. For both sets of experiments, the encodings proposed in Section~\ref{sec:method} were used.}
\label{fig:intermediate}
\end{figure*}

\subsection*{Performance Predictors Analysis}

In the first ablation study, the predictor models were compared by keeping the training set size fixed at 300 and using the encoding methods proposed in Section~\ref{sec:method} to generate input features. The performance of the surrogate models were evaluated using correlation values as the reference metric. These correlation values were obtained by analysing sub-networks extracted from the available configurations of the OFAv2 algorithm, which also includes sub-networks from OFA.
The aim of this study was to identify the most accurate predictor of sub-network accuracy that could be used in both NAT and NATv2 evaluations. To obtain the most accurate estimate, all models were evaluated using the 10-fold cross-validation technique. After calculating the validation performance by averaging, the 10 models trained on different data splits were aggregated by ensemble to form a single macro-model. The best macro-model among these ensembles was selected as the predictor.

Referring to the candidate models presented in Section~\ref{sec:method}, Figure~\ref{fig:predictor_by_network} shows their performance, measured as the Spearman correlation between the predicted accuracy and the actual accuracy, referred to as the ``rho correlation". The best performing models were CatBoost and LGBM, both above 0.9, followed by CARTS and E2EPP. CatBoost and LGBM also consistently produced the smallest error standard deviation intervals.
This result has two positive implications. Firstly, it demonstrates the correlation between the proposed encoding and the accuracy of the corresponding model. Secondly, it means that the encoding successfully captures the large heterogeneity of possible networks, resulting in excellent error standard deviation intervals for the metric under consideration.

Figure~\ref{fig:time_predictor_by_network} illustrates the time required by the different models to complete the entire training process. Given the reasonable size of the dataset, it is not surprising that the average times are generally positive for most models. As expected, the CART, RIDGE and BAYESIAN models are the fastest due to their simplicity. However, as their performance was not so promising, these models were not considered.
When choosing a performance predictor, maximising the regression metric (in this case the rho correlation) is a crucial consideration. However, it is not the only factor to consider. The time taken to fit each model is also important. Although CatBoost performed slightly better than LGBM, it was significantly slower to fit. As a result, LGBM was ultimately selected as the predictor model.

Once the predictor model was fixed, the correlation values were evaluated for different training set sizes, using both integer and one-hot encodings to generate the input features. The results shown in Figure~\ref{fig:predictor_by_encoding} indicate that for the given configuration, integer encoding outperforms one-hot encoding in terms of performance and stability. Furthermore, it appears that the number of samples reaches an optimal point, as the improvements in rho correlation seem to plateau in both encoding configurations.
Finally, considering the performance variance achieved with smaller datasets, it can be concluded that the decision to replace the small but growing archive of NAT with a larger fixed-size archive in NATv2 is advantageous and improves the accuracy of the best estimated sub-networks from the beginning of the search process.

\subsection*{OFAv2 Super-Networks Analysis}

The second ablation study focuses on evaluating the effectiveness of the NAT algorithm enhanced with the new encoding methods described in Section~\ref{sec:method} and the new performance predictor. This study serves as a preliminary evaluation to determine the improvements achieved by our approach compared to the initial model at this stage. The aim of this analysis is to evaluate the variation in performance resulting from the modification of the initial super-network. Specifically, two variations are considered: the initial super-network obtained via OFA and the starting super-networks obtained via OFAv2. 

The results of this study are summarised in Figure~\ref{fig:intermediate}. Each row corresponds to a different dataset and each column represents a different optimisation objective. Starting from the first row, which corresponds to the experiments carried out on CIFAR10, it can be seen that changing the super-network leads to an overall improvement in all the configurations analysed.
When optimising for accuracy only, the best model obtained from OFAv2 shows an improvement in accuracy of about 2\% compared to the model obtained from OFA. Moving on to the results of the multi-objective searches, it can be observed that the models obtained not only have higher accuracy, but also have significantly fewer parameters and MACs on average. For example, when comparing architectures with fewer parameters, in addition to a 1\% accuracy advantage for the architecture found by OFAv2, the number of parameters is approximately five times lower compared to the best architecture found by OFA.

Looking at the results obtained on CIFAR100 and Tiny ImageNet, which are more complex datasets compared to CIFAR10, the previous observations are even more significant. In particular, when focusing only on accuracy optimisations, there is an improvement of about 5\% and 9\% respectively in favour of the models obtained from OFAv2. The performance of the solutions found in multi-objective optimisations further supports the claim that the proposed approach becomes more effective as the complexity of the problem increases.
In addition, these results highlight the success and effectiveness of the proposed encoding method in this context. The encoding demonstrates its ability to capture the complexity of the problem and contribute to the improved performance of the models.

\subsection*{Post-Processing Optimisation}

The third and final ablation study focuses on fine-tuning the parameters of the post-processing strategy once the optimal sub-networks have been found by NATv2. The goal of this study is to determine the optimal combination of optimiser, initial learning rate, and weight assignment for networks with early exits. In the case of networks with early exits, the choice of weights applied to the outputs is also part of the tuning process. The results presented in this study are based on the Tiny ImageNet dataset, with NATv2 run using the ``Accuracy \& Params" and ``Accuracy \& MACs" objectives.

We found that for each optimiser, regardless of the type of post-processing, the average performance obtained by setting the initial learning rate to $10^{-4}$ was better than that obtained by setting it to $10^{-5}$. This finding is consistent with the expectation that starting with a low learning rate may result in too much dilution of learning due to the cosine annealing schedule.

Regarding the choice of optimizer, single-exit architectures tend to benefit from using SGD, while almost all multiple-exit architectures show better performance with AdamW and Ranger optimizers, with a slightly stronger preference for AdamW.
In terms of the AEP strategy, it was found that using a uniform weight distribution for networks with early exits yields better results on average.

Quantitatively, as demonstrated by the final results of this work, presented below in Table~\ref{table:final}, the inclusion of post-processing provides significant benefits to model accuracy, with an average improvement of 2.63\% on Tiny ImageNet. However, it is important to consider the trade-off in terms of increased parameters (21.84\% increase) and MACs (7.63\% increase) when early exits are incorporated. Nevertheless, the architectures discovered by NATv2 often exhibit high computational efficiency, and the increase in complexity can be justified by the significant performance gains. Additionally, it is worth noting that the post-processing step is optional and that incurs a minimal time cost of a few minutes, which is negligible compared to the overall research process.

\begin{table*}[!t]
\centering
\caption{The results extracted from the final set of experiments. For each dataset, the first column on the left shows the best sub-networks found, grouped by research objective. For each dataset and metric, the best result is highlighted in \textbf{bold}. For each dataset, metric and objective, the best result is \underline{underlined}. For each multi-objective optimisation, the best model found for accuracy, and the best model for the second objective of the optimisation are reported. Each NATv2 experiment is also presented in its post-processed form called NATv2 + PP.}
\begin{tabular}{|c|llccc|}
\hline
& \textbf{Optimisation Objective} & \textbf{Model} & \textbf{Accuracy} & \textbf{Params (M)} & \textbf{MACs (M)}\\
\hline
\hline
 \multirow{15}{*}{\rotatebox[origin=c]{90}{\textbf{CIFAR10}}} & & NAT (Baseline) & 90.68 & \underline{6.75} & \underline{59.97}\\
 & Accuracy & NATv2 & 93.06 & 8.74 & 65.74\\
 & & NATv2 + PP & \underline{\textbf{93.17}} & 12.42 & 77.82\\
\cline{2-6}
 & & NAT (Baseline) & 90.73 & 2.91 & 30.03\\
 & & NAT (Baseline) & 90.65 & 2.51 & 26.74\\
 & & NATv2 & 92.69 & 1.30 & 34.39\\
 & \smash{\raisebox{3pt}{Accuracy \& Params}} & NATv2 & 91.46 & \underline{0.27} & \underline{12.52}\\
 & & NATv2 + PP & \underline{93.06} & 1.56 & 37.75\\
 & & NATv2 + PP & 92.00 & 0.47 & 24.28\\
\cline{2-6}
 & & NAT (Baseline) & 89.98 & 2.40 & 15.33\\
 & & NAT (Baseline) & 85.66 & 2.14 & 7.82\\
 & & NATv2 & 91.77 & 1.09 & 20.11\\
 & \smash{\raisebox{3pt}{Accuracy \& MACs}} & NATv2 & 89.67 & \underline{\textbf{0.19}} & \underline{\textbf{6.35}}\\
 & & NATv2 + PP & \underline{92.29} & 1.35 & 22.75\\
 & & NATv2 + PP & 90.23 & 0.23 & 6.92\\
\hline
\hline
\multirow{15}{*}{\rotatebox[origin=c]{90}{\textbf{CIFAR100}}} & & NAT (Baseline) & 66.93 & \underline{6.26} & \underline{55.72}\\
 & Accuracy & NATv2 & 71.88 & 9.75 & 56.60\\
 & & NATv2 + PP & \underline{\textbf{73.39}} & 11.13 & 74.28\\
\cline{2-6}
 & & NAT (Baseline) & 66.83 & 3.62 & 31.44\\
 & & NAT (Baseline) & 65.56 & 2.57 & 26.69\\
 & & NATv2 & 70.68 & 1.36 & 34.26\\
 & \smash{\raisebox{3pt}{Accuracy \& Params}} & NATv2 & 69.50 & \underline{0.86} & \underline{21.62}\\
 & & NATv2 + PP & \underline{72.03} & 1.70 & 31.79\\
 & & NATv2 + PP & 70.54 & 1.03 & 23.38\\
\cline{2-6}
 & & NAT (Baseline) & 64.76 & 2.70 & 16.05\\
 & & NAT (Baseline) & 58.61 & 2.26 & 7.94\\
 & & NATv2 & 69.31 & 1.26 & 21.34\\
 & \smash{\raisebox{3pt}{Accuracy \& MACs}} & NATv2 & 66.29 & \underline{\textbf{0.21}} & \underline{\textbf{8.14}}\\
 & & NATv2 + PP & \underline{71.02} & 1.59 & 24.04\\
 & & NATv2 + PP & 67.90 & 0.26 & 8.93\\
\hline
\hline
\multirow{15}{*}{\rotatebox[origin=c]{90}{\textbf{Tiny ImageNet}}} & & NAT (Baseline) & 43.06 & 8.10 & 61.67\\
 & Accuracy & NATv2 & 53.59 & \underline{1.66} & \underline{43.43}\\
 & & NATv2 + PP & \underline{\textbf{54.82}} & 2.06 & 46.94\\
\cline{2-6}
 & & NAT (Baseline) & 43.45 & 4.05 & 46.80\\
 & & NAT (Baseline) & 42.99 & 2.71 & 28.00\\
 & & NATv2 & 51.16 & 1.44 & 39.19\\
 & \smash{\raisebox{3pt}{Accuracy \& Params}} & NATv2 & 39.92 & \underline{\textbf{0.10}} & \underline{\textbf{5.43}}\\
 & & NATv2 + PP & \underline{54.31} & 1.85 & 41.70\\
 & & NATv2 + PP & 45.03 & \underline{\textbf{0.10}} & \underline{\textbf{5.43}}\\
\cline{2-6}
 & & NAT (Baseline) & 42.00 & 2.86 & 17.01\\
 & & NAT (Baseline) & 38.89 & 2.39 & 8.06\\
 & & NATv2 & 51.05 & 1.46 & 28.41\\
 & \smash{\raisebox{3pt}{Accuracy \& MACs}} & NATv2 & 47.24 & \underline{0.25} & \underline{5.95}\\
 & & NATv2 + PP & \underline{53.91} & 1.87 & 31.97\\
 & & NATv2 + PP & 48.96 & 0.32 & 6.55\\
\hline
\end{tabular}
\label{table:final}
\end{table*}

\subsection*{Final Results}

The final set of experiments aims to compare the performance of the NAT reference model and its enriched version, NATv2, with and without the post-processing step. The results of these experiments are presented in Table~\ref{table:final}, which encompasses the evaluation on CIFAR10, CIFAR100, and Tiny ImageNet datasets. The results are reported in terms of top-1 accuracy, number of parameters, and number of multiply-accumulate (MAC) operations, both measured in millions.

% CIFAR10
For CIFAR10, the NATv2 + PP model obtaines the highest accuracy of 93.17\%, which is better than the other models. 
However, it also has the highest number of parameters (12.42M) and MAC (77.82M). 
Considering the trade-off between accuracy and number of parameters, the NATv2 model obtained by optimising ``Accuracy \ Params'' with only 0.27M parameters achieved an accuracy of 91.46\%, which is higher than any model obtained using NAT. 
Similarly, considering the trade-off between accuracy and MACs, the NATv2 model obtained by the ``Accuracy \ MACs'' optimisation achieves an accuracy of 89.67\% with a minimum number of MACs equal to 6.35M and a very small number of parameters equal to 0.19M. This is particularly suitable for devices with very strict constraints.
Generally speaking, it is fair to say that for each optimisation scenario considered, NATv2 significantly outperforms NAT for each of the research objectives in most cases. 
On average, without exceeding either accuracy or parameter gains, the application of post-processing is beneficial.

% CIFAR100
If we turn to the CIFAR100 dataset, we observe similar trends in terms of the performance of the models and the optimisation targets.
The NATv2 + PP model found in the single-target search emerges as the best performer overall, achieving an accuracy of 73.39\%. This result demonstrates the model's ability to cope with the increased complexity of the CIFAR100 dataset, far exceeding the baseline accuracy of 66.93\%.
Considering the trade-off between accuracy and parameter count, NATv2 successfully optimises the search, finding a model with 0.86M parameters and an accuracy of 69.50\%. Compared to the NAT models identified by the same optimisation, this is a significant improvement.
It is also interesting to note that this model found with NATv2, although in a multi-objective optimisation, performs better in terms of accuracy than even the best NAT model optimised without taking parameters and MACs into account. It improves its accuracy by 2.57\% and reduces parameters and MACs by 86.26\% and 38.80\% respectively.
The NATv2 model stands out again when considering the trade-off between accuracy and MAC. 
The model with the lowest number of MACs (8.14M) consists of a very small number of parameters, 0.21M, with an accuracy of 66.29\%, which is higher or comparable to any model found by NAT in any optimisation configuration. In addition to losing 7.68 percentage points of accuracy, NAT's lightest model requires more than twelve times as many parameters and twice as many MACs.

% TINY
Let us now turn our attention to the results obtained for the Tiny ImageNet data set, which represents the most challenging problem.
Considering accuracy as the only optimisation objective, the NATv2 + PP model achieves the highest accuracy of 54.82\%, which exceeds the baseline accuracy of 43.06\%. This is a further evidence of the effectiveness of the NATv2 + PP model in the improvement of classification performance.
In the analysis of the trade-off between accuracy and number of parameters, NATv2 results in a model with only 0.10 million parameters, but still achieves an accuracy of 39.92%.
By applying post-processing to this model, which is obviously an architecture without the possibility of inserting early exits and therefore extremely flat, the accuracy rises to 45.03\% without increasing either the parameters or the MACs. This further demonstrates the usefulness of this fine-tuning step. The model thus obtained turns out to be better than any model found by NAT, this time with a truly minimal number of parameters.

% GENERALE
Overall, the experiments show that NATv2 + PP consistently achieves the highest level of accuracy across all three data sets. However, by achieving competitive accuracy with significantly fewer parameters, NATv2 demonstrates its strength in terms of parameter efficiency. Furthermore, by achieving reasonable accuracy and minimising computational requirements, NATv2 also demonstrates its efficiency in terms of MAC.
In general, it can be said that NATv2 is able to achieve significantly better trade-offs than NAT, in some cases decimating the number of parameters, and is therefore particularly suitable for model searches for devices with a small amount of memory.
On the other hand, if secondary objectives are to be sacrificed for the sake of accuracy, the use of post-processing has proven to be a beneficial step in the vast majority of cases, as it allows for architectures that are still considerably lighter and faster than those realised by NAT, while at the same time achieving much better performance in terms of accuracy.

\section{Conclusion}\label{sec:conclusion}

In this paper we have presented Neural Architecture Transfer 2 (NATv2), the extension of the Neural Architecture Transfer (NAT) technique by the implementation of two recent algorithms, namely Once-For-All-2 (OFAv2) and Anticipate Ensemble and Prune (AEP).
In particular, we have shown that NATv2 can find networks that are significantly smaller in terms of parameters and operations, and more accurate than those found by applying NAT, by modifying the architectural design of the super-networks as well as the algorithm used. 

The greatest improvements were obtained by applying NATv2 exactly to these modified super-networks. Among the modifications, the one that contributed to the greatest improvements was the introduction of early exits in the architectures.
Among the most important algorithmic improvement, the introduction of the post-processing phase, which allows further refinement of the returned sub-networks, proved to be an extremely effective addition, virtuously increasing the performance at a negligible cost in terms of parameters and operations.
The results suggest that NATv2 is a successful extension of NAT, which was already an excellent tool for the realisation of deep learning models by satisfying very strict constraints. In particular, NATv2 is highly recommended for exploring high-performance architectures with an extremely small number of parameters.

\section*{Acknowledgment}\label{sec:ack}

This project has been supported by AI-SPRINT: AI in Secure Privacy-pReserving computINg conTinuum (European Union H2020 grant agreement No. 101016577) and FAIR: Future Artificial Intelligence Research (NextGenerationEU, PNRR-PE-AI scheme, M4C2, investment 1.3, line on Artificial Intelligence).

%Bibliography
\bibliographystyle{unsrt}  
\bibliography{references}

\begin{thebibliography}{10}

\bibitem{liu2022convnet}
Zhuang Liu, Hanzi Mao, Chao-Yuan Wu, Christoph Feichtenhofer, Trevor Darrell,
  and Saining Xie.
\newblock A convnet for the 2020s.
\newblock In {\em Proceedings of the IEEE/CVF Conference on Computer Vision and
  Pattern Recognition}, pages 11976--11986, 2022.

\bibitem{desislavov2021compute}
Radosvet Desislavov, Fernando Mart{\'\i}nez-Plumed, and Jos{\'e}
  Hern{\'a}ndez-Orallo.
\newblock Compute and energy consumption trends in deep learning inference.
\newblock {\em arXiv preprint arXiv:2109.05472}, 2021.

\bibitem{zoph2018learning}
Barret Zoph, Vijay Vasudevan, Jonathon Shlens, and Quoc~V. Le.
\newblock Learning transferable architectures for scalable image recognition,
  2018.

\bibitem{liu2018darts}
Hanxiao Liu, Karen Simonyan, and Yiming Yang.
\newblock Darts: Differentiable architecture search.
\newblock {\em arXiv preprint arXiv:1806.09055}, 2018.

\bibitem{cai2019once}
Han Cai, Chuang Gan, Tianzhe Wang, Zhekai Zhang, and Song Han.
\newblock Once-for-all: Train one network and specialize it for efficient
  deployment.
\newblock {\em arXiv preprint arXiv:1908.09791}, 2019.

\bibitem{sarti2023enhancing}
Simone Sarti, Eugenio Lomurno, Andrea Falanti, and Matteo Matteucci.
\newblock Enhancing once-for-all: A study on parallel blocks, skip connections
  and early exits.
\newblock {\em arXiv preprint arXiv:2302.01888}, 2023.

\bibitem{lu2021neural}
Zhichao Lu, Gautam Sreekumar, Erik Goodman, Wolfgang Banzhaf, Kalyanmoy Deb,
  and Vishnu~Naresh Boddeti.
\newblock Neural architecture transfer.
\newblock {\em IEEE Transactions on Pattern Analysis and Machine Intelligence},
  43(9):2971--2989, 2021.

\bibitem{elsken2019neural}
Thomas Elsken, Jan~Hendrik Metzen, and Frank Hutter.
\newblock Neural architecture search: A survey.
\newblock {\em The Journal of Machine Learning Research}, 20(1):1997--2017,
  2019.

\bibitem{he2016deep}
Kaiming He, Xiangyu Zhang, Shaoqing Ren, and Jian Sun.
\newblock Deep residual learning for image recognition.
\newblock In {\em Proceedings of the IEEE conference on computer vision and
  pattern recognition}, pages 770--778, 2016.

\bibitem{szegedy2015going}
Christian Szegedy, Wei Liu, Yangqing Jia, Pierre Sermanet, Scott Reed, Dragomir
  Anguelov, Dumitru Erhan, Vincent Vanhoucke, and Andrew Rabinovich.
\newblock Going deeper with convolutions.
\newblock In {\em Proceedings of the IEEE conference on computer vision and
  pattern recognition}, pages 1--9, 2015.

\bibitem{liu2018progressive}
Chenxi Liu, Barret Zoph, Maxim Neumann, Jonathon Shlens, Wei Hua, Li-Jia Li,
  Li~Fei-Fei, Alan Yuille, Jonathan Huang, and Kevin Murphy.
\newblock Progressive neural architecture search.
\newblock In {\em Proceedings of the European conference on computer vision
  (ECCV)}, pages 19--34, 2018.

\bibitem{lomurno2021pareto}
Eugenio Lomurno, Stefano Samele, Matteo Matteucci, and Danilo Ardagna.
\newblock Pareto-optimal progressive neural architecture search.
\newblock In {\em Proceedings of the Genetic and Evolutionary Computation
  Conference Companion}, pages 1726--1734, 2021.

\bibitem{falanti2022popnasv2}
Andrea Falanti, Eugenio Lomurno, Stefano Samele, Danilo Ardagna, and Matteo
  Matteucci.
\newblock Popnasv2: An efficient multi-objective neural architecture search
  technique.
\newblock In {\em 2022 International Joint Conference on Neural Networks
  (IJCNN)}, pages 1--8. IEEE, 2022.

\bibitem{falanti2023popnasv3}
Andrea Falanti, Eugenio Lomurno, Danilo Ardagna, and Matteo Matteucci.
\newblock Popnasv3: A pareto-optimal neural architecture search solution for
  image and time series classification.
\newblock {\em Applied Soft Computing}, page 110555, 2023.

\bibitem{real2019regularized}
Esteban Real, Alok Aggarwal, Yanping Huang, and Quoc~V Le.
\newblock Regularized evolution for image classifier architecture search.
\newblock In {\em Proceedings of the aaai conference on artificial
  intelligence}, volume~33, pages 4780--4789, 2019.

\bibitem{lu2019nsga}
Zhichao Lu, Ian Whalen, Vishnu Boddeti, Yashesh Dhebar, Kalyanmoy Deb, Erik
  Goodman, and Wolfgang Banzhaf.
\newblock Nsga-net: neural architecture search using multi-objective genetic
  algorithm.
\newblock In {\em Proceedings of the genetic and evolutionary computation
  conference}, pages 419--427, 2019.

\bibitem{bender2018understanding}
Gabriel Bender, Pieter-Jan Kindermans, Barret Zoph, Vijay Vasudevan, and Quoc
  Le.
\newblock Understanding and simplifying one-shot architecture search.
\newblock In {\em International conference on machine learning}, pages
  550--559. PMLR, 2018.

\bibitem{guo2020single}
Zichao Guo, Xiangyu Zhang, Haoyuan Mu, Wen Heng, Zechun Liu, Yichen Wei, and
  Jian Sun.
\newblock Single path one-shot neural architecture search with uniform
  sampling.
\newblock In {\em Computer Vision--ECCV 2020: 16th European Conference,
  Glasgow, UK, August 23--28, 2020, Proceedings, Part XVI 16}, pages 544--560.
  Springer, 2020.

\bibitem{hinton2015distilling}
Geoffrey Hinton, Oriol Vinyals, Jeff Dean, et~al.
\newblock Distilling the knowledge in a neural network.
\newblock {\em arXiv preprint arXiv:1503.02531}, 2(7), 2015.

\bibitem{benyahia2019overcoming}
Yassine Benyahia, Kaicheng Yu, Kamil~Bennani Smires, Martin Jaggi, Anthony~C
  Davison, Mathieu Salzmann, and Claudiu Musat.
\newblock Overcoming multi-model forgetting.
\newblock In {\em International Conference on Machine Learning}, pages
  594--603. PMLR, 2019.

\bibitem{sandler2018mobilenetv2}
Mark Sandler, Andrew Howard, Menglong Zhu, Andrey Zhmoginov, and Liang-Chieh
  Chen.
\newblock Mobilenetv2: Inverted residuals and linear bottlenecks.
\newblock In {\em Proceedings of the IEEE conference on computer vision and
  pattern recognition}, pages 4510--4520, 2018.

\bibitem{howard2019searching}
Andrew Howard, Mark Sandler, Grace Chu, Liang-Chieh Chen, Bo~Chen, Mingxing
  Tan, Weijun Wang, Yukun Zhu, Ruoming Pang, Vijay Vasudevan, et~al.
\newblock Searching for mobilenetv3.
\newblock In {\em Proceedings of the IEEE/CVF international conference on
  computer vision}, pages 1314--1324, 2019.

\bibitem{sarti2023anticipate}
Simone Sarti, Eugenio Lomurno, and Matteo Matteucci.
\newblock Anticipate, ensemble and prune: Improving convolutional neural
  networks via aggregated early exits.
\newblock {\em arXiv preprint arXiv:2301.12168}, 2023.

\bibitem{williams1995gaussian}
Christopher Williams and Carl Rasmussen.
\newblock Gaussian processes for regression.
\newblock {\em Advances in neural information processing systems}, 8, 1995.

\bibitem{bors2001introduction}
Adrian~G Bors.
\newblock Introduction of the radial basis function (rbf) networks.
\newblock In {\em Online symposium for electronics engineers}, 2001.

\bibitem{rumelhart1985learning}
David~E Rumelhart, Geoffrey~E Hinton, and Ronald~J Williams.
\newblock Learning internal representations by error propagation.
\newblock Technical report, California Univ San Diego La Jolla Inst for
  Cognitive Science, 1985.

\bibitem{loh2011classification}
Wei-Yin Loh.
\newblock Classification and regression trees.
\newblock {\em Wiley interdisciplinary reviews: data mining and knowledge
  discovery}, 1(1):14--23, 2011.

\bibitem{drucker1996support}
Harris Drucker, Christopher~J Burges, Linda Kaufman, Alex Smola, and Vladimir
  Vapnik.
\newblock Support vector regression machines.
\newblock {\em Advances in neural information processing systems}, 9, 1996.

\bibitem{hoerl1970ridge}
Arthur~E Hoerl and Robert~W Kennard.
\newblock Ridge regression: Biased estimation for nonorthogonal problems.
\newblock {\em Technometrics}, 12(1):55--67, 1970.

\bibitem{fix1989discriminatory}
Evelyn Fix and Joseph~Lawson Hodges.
\newblock Discriminatory analysis. nonparametric discrimination: Consistency
  properties.
\newblock {\em International Statistical Review/Revue Internationale de
  Statistique}, 57(3):238--247, 1989.

\bibitem{tipping2001sparse}
Michael~E Tipping.
\newblock Sparse bayesian learning and the relevance vector machine.
\newblock {\em Journal of machine learning research}, 1(Jun):211--244, 2001.

\bibitem{sun2019surrogate}
Yanan Sun, Handing Wang, Bing Xue, Yaochu Jin, Gary~G Yen, and Mengjie Zhang.
\newblock Surrogate-assisted evolutionary deep learning using an end-to-end
  random forest-based performance predictor.
\newblock {\em IEEE Transactions on Evolutionary Computation}, 24(2):350--364,
  2019.

\bibitem{ke2017lightgbm}
Guolin Ke, Qi~Meng, Thomas Finley, Taifeng Wang, Wei Chen, Weidong Ma, Qiwei
  Ye, and Tie-Yan Liu.
\newblock Lightgbm: A highly efficient gradient boosting decision tree.
\newblock {\em Advances in neural information processing systems}, 30, 2017.

\bibitem{prokhorenkova2018catboost}
Liudmila Prokhorenkova, Gleb Gusev, Aleksandr Vorobev, Anna~Veronika Dorogush,
  and Andrey Gulin.
\newblock Catboost: unbiased boosting with categorical features.
\newblock {\em Advances in neural information processing systems}, 31, 2018.

\bibitem{le2015tiny}
Ya~Le and Xuan Yang.
\newblock Tiny imagenet visual recognition challenge.
\newblock {\em CS 231N}, 7(7):3, 2015.

\bibitem{krizhevsky2009learning}
Alex Krizhevsky, Geoffrey Hinton, et~al.
\newblock Learning multiple layers of features from tiny images.
\newblock 2009.

\bibitem{loshchilov2017decoupled}
Ilya Loshchilov and Frank Hutter.
\newblock Decoupled weight decay regularization.
\newblock {\em arXiv preprint arXiv:1711.05101}, 2017.

\bibitem{zhang2019lookahead}
Michael Zhang, James Lucas, Jimmy Ba, and Geoffrey~E Hinton.
\newblock Lookahead optimizer: k steps forward, 1 step back.
\newblock {\em Advances in neural information processing systems}, 32, 2019.

\bibitem{liu2019variance}
Liyuan Liu, Haoming Jiang, Pengcheng He, Weizhu Chen, Xiaodong Liu, Jianfeng
  Gao, and Jiawei Han.
\newblock On the variance of the adaptive learning rate and beyond.
\newblock {\em arXiv preprint arXiv:1908.03265}, 2019.

\bibitem{blank2019investigating}
Julian Blank, Kalyanmoy Deb, and Proteek~Chandan Roy.
\newblock Investigating the normalization procedure of nsga-iii.
\newblock In {\em Evolutionary Multi-Criterion Optimization: 10th International
  Conference, EMO 2019, East Lansing, MI, USA, March 10-13, 2019, Proceedings
  10}, pages 229--240. Springer, 2019.

\bibitem{blank2020pymoo}
Julian Blank and Kalyanmoy Deb.
\newblock Pymoo: Multi-objective optimization in python.
\newblock {\em IEEE Access}, 8:89497--89509, 2020.

\end{thebibliography}

\end{document}